\pdfoutput=1

\documentclass[11pt]{article}

\usepackage[]{acl2021}

\usepackage{times}
\usepackage{latexsym}

\usepackage[T1]{fontenc}

\usepackage[utf8]{inputenc}

\usepackage{microtype}

\usepackage{makecell}

\usepackage{microtype}

\usepackage{graphicx}
\usepackage{paralist}
\usepackage{url}
\usepackage{xspace}
\usepackage{amsfonts,amsmath}
\usepackage{booktabs}
\usepackage{setspace}
\usepackage{multirow}
\usepackage{multicol}
\usepackage{subfigure}
\usepackage{hyperref}
\usepackage{xcolor}
\usepackage{amssymb}
\usepackage{pifont}
\usepackage{color, colortbl}
\usepackage{bbm}
\usepackage{wrapfig}
\usepackage{lipsum}
\usepackage{enumitem}
\usepackage{paralist, tabularx}
\usepackage{balance}

\newcommand*\colourTrue[1]{%
  \expandafter\newcommand\csname #1True\endcsname{\textcolor{#1}{\ding{52}}}%
}
\newcommand{\model}{\textsc{ReFeed} }
\colourTrue{teal}
\colourTrue{red}

\newcommand*\colourFalse[1]{%
  \expandafter\newcommand\csname #1False\endcsname{\textcolor{#1}{\ding{56}}}%
}
\colourFalse{green}
\colourFalse{red}

\title{Improving Language Models via Plug-and-Play Retrieval Feedback}

\author{
Wenhao Yu$^{\clubsuit}$, 
Zhihan Zhang$^{\clubsuit}$, Zhenwen Liang$^{\clubsuit}$, \\ \textbf{Meng Jiang$^{\clubsuit}$, Ashish Sabharwal$^{\spadesuit}$} \\
$^{\clubsuit}$University of Notre Dame;  $^{\spadesuit}$Allen Institute for Artificial Intelligence  \\
{$^{\clubsuit}$\tt wyu1@nd.edu;}  {$^{\spadesuit}$\tt ashishs@allenai.org}
}

\begin{document}
\maketitle

\begin{abstract}

Large language models (LLMs) exhibit remarkable performance across various NLP tasks. However, they often generate incorrect or hallucinated information, which hinders their practical applicability in real-world scenarios. Human feedback has been shown to effectively enhance the factuality and quality of generated content, addressing some of these limitations. However, this approach is resource-intensive, involving manual input and supervision, which can be time-consuming and expensive. Moreover, it cannot be provided during inference, further limiting its practical utility in dynamic and interactive applications. In this paper, we introduce \textsc{ReFeed}, a novel pipeline designed to enhance LLMs by providing automatic retrieval feedback in a plug-and-play framework without the need for expensive fine-tuning. \textsc{ReFeed} first generates initial outputs, then utilizes a retrieval model to acquire relevant information from large document collections, and finally incorporates the retrieved information into the in-context demonstration for output refinement, thereby addressing the limitations of LLMs in a more efficient and cost-effective manner.
Experiments on four knowledge-intensive benchmark datasets demonstrate our proposed \textsc{ReFeed} could improve over +6.0\% under zero-shot setting and +2.5\% under few-shot setting, compared to baselines without using retrieval feedback.

\end{abstract}

\section{Introduction}
Large language models (LLMs) have demonstrated exceptional performance in various NLP tasks, utilizing in-context learning to eliminate the need for task-specific fine-tuning~\citep{brown2020language,chowdhery2022palm,openai2023gpt4}. Such models are typically trained on extensive datasets, capturing a wealth of world or domain-specific knowledge within their parameters.

Despite these achievements, LLMs exhibit certain shortcomings, particularly when confronted with complex reasoning and knowledge-intensive tasks~\citep{zhang2023automatic,yu2023generate}. 
One prominent drawback is their propensity to hallucinate content, generating information not grounded by world knowledge, leading to untrustworthy outputs and a diminished capacity to provide accurate information~\citep{yu2022survey,manakul2023selfcheckgpt,alkaissi2023artificial}.
Another limitation of LLMs is the quality and scope of the knowledge they store. The knowledge embedded within an LLM may be incomplete or out-of-date due to the reliability of the sources in the pre-training corpus~\citep{lazaridou2022internet,shi2023replug}. The vastness of the information landscape exacerbates this issue, making it difficult for models to maintain a comprehensive and up-to-date understanding of world facts.
Moreover, LLMs cannot ``memorize'' all world information, especially struggling with the long tail of knowledge from their training corpus~\citep{mallen2022not,kandpal2022large}. This inherent limitation compels them to balance the storage and retrieval of diverse and rare knowledge against focusing on more frequently encountered information, leading to potential inaccuracies when addressing questions related to less common topics or requiring nuanced understanding.

Existing methods for enhancing the factuality of language models involve adjusting model outputs based on human feedback, followed by reinforcement learning-based fine-tuning~\citep{nakano2021webgpt,campos2022training,ouyang2022training,liu2023languages}. While this approach simulates human-to-human task learning environments, fine-tuning LLMs, can be exceedingly costly due to the exponential growth in LLM size and the necessity for annotators to provide extensive feedback. 
Over-reliance on positively-rated data may limit the model's ability to identify and rectify negative attributes or errors, potentially hampering its capacity to generalize to unseen data and novel scenarios.
Furthermore, once LLMs are fine-tuned, they are unable to receive real-time feedback during inference or facilitate immediate error correction.

In this paper, we aim to provide automatic feedback in a plug-and-play manner without the need for fine-tuning LLMs. We explore two primary research questions: First, can we employ a retrieval method to provide feedback on individual generated outputs without relying on human annotators?
Second, can we integrate the feedback to refine previous outputs in a plug-and-play manner, circumventing the expensive fine-tuning of language models? With regards to the two questions posed, we propose a novel pipeline for improving language model inference through automatic retrieval feedback, named \textsc{ReFeed}, in a plug-and-play framework. Specifically, the language model generates initial outputs, followed by a retrieval model using the original query and generated outputs as a new query to retrieve relevant information from large document collections like Wikipedia. The retrieved information enables the language model to reconsider the generated outputs and refine them, potentially producing a new output (though it may remain the same if no changes are made).

Notably, compared to retrieve-then-read methods~\citep{lewis2020retrieval,lazaridou2022internet,shi2023replug}, our method benefits from more relevant documents retrieved from the corpus to directly elucidate the relationship between query and outputs. Besides, without generating the initial output, the supporting document cannot be easily retrieved due to the lack of lexical or semantic overlap with the question. 
We discuss the comparison in more detail in related work and experiments.

To further enhance our proposed \textsc{ReFeed} pipeline, we introduce two innovative modules within this framework. Firstly, we diversify the initial generation step to produce multiple output candidates, enabling the model to determine the most reliable answer by analyzing the diverse set of retrieved documents. Secondly, we adopt an ensemble approach that merges language model outputs before and after retrieval feedback using a perplexity ranking method, as the retrieval feedback may occasionally mislead the language model (refer to the case study in Figure \ref{fig:case} for details).

Overall, our main contributions can be summarized as follows:
\begin{enumerate}[wide=10pt, itemsep=-0.2ex, topsep=-5pt,]
    \item We propose a novel pipeline utilizing retrieval feedback, named \textsc{ReFeed} to improve large language models in a plug-and-play manner.
    \item We design two novel modules to advance the \textsc{ReFeed} pipeline, specifically diversifying generation outputs and ensembling initial and post-feedback answers.
    \item Our experiments on three challenging knowledge-intensive tasks demonstrate that \textsc{ReFeed} can achieve state-of-the-art performance under the few-shot setting.
\end{enumerate}

\section{Related Work}
\subsection{Solving Knowledge-intensive Tasks via Retrieve-then-Read Pipeline}

Mainstream methods for solving knowledge-intensive NLP tasks employ a \textit{retrieve-then-read} model pipeline. Given an input query, a retriever is employed to search a large evidence corpus (e.g., Wikipedia) for relevant documents that may contain the answer. Subsequently, a reader is used to scrutinize the retrieved documents and predict an answer. Recent research has primarily focused on improving either the retriever~\citep{karpukhin2020dense,qu2021rocketqa,sachan2022questions} or the reader~\citep{izacard2021leveraging, yu2022kg,ju2022grape}, or training the entire system end-to-end~\citep{singh2021end,shi2023replug}.
Early retrieval methods largely employed sparse retrievers, such as BM25~\citep{chen2017reading}. Recently, ORQA~\citep{lee2019latent} and DPR \citep{karpukhin2020dense} have revolutionized the field by using dense contextualized vectors for document indexing, resulting in superior performance compared to traditional approaches. 
Recently, several work proposed to replace the retrieval model with a large language model as retriever, owing to the powerful knowledge memorization capabilties~\citep{yu2023generate,sun2023recitation}. However, these methods may still be prone to hallucination issues and are unable to access up-to-date information. 
Notably, compared to \textit{retrieve-then-read} pipelines like RePLUG~\citep{shi2023replug}, our method benefits from more relevant documents retrieved from the corpus to directly elucidate the relationship between query and outputs. Additionally, without generating the initial output, the text supporting the output cannot be easily identified due to the lack of lexical or semantic overlap with the question.


\subsection{Aligning Language Model with Instructions via Human Feedback}

Human feedback plays a crucial role in evaluating language model performance, addressing accuracy, fairness, and bias issues, and offering insights for model improvement to better align with human expectations. Recognizing the significance of integrating human feedback into language models, researchers have developed and tested various human-in-the-loop methodologies~\citep{nakano2021webgpt, campos2022training, ouyang2022training, liu2023languages, scheurer2023training}. 
InstructGPT~\citep{ouyang2022training} was a trailblazer in this domain, utilizing reinforcement learning from human feedback (RLHF) to fine-tune GPT-3 to adhere to a wide range of written instructions. It trained a reward model (RM) on this dataset to predict the preferred model output based on human labelers' preferences. The RM then served as a reward function, and the supervised learning baseline was fine-tuned to maximize this reward using the PPO algorithm.
\citet{liu2023languages} proposed converting all forms of feedback into sentences, which were subsequently used to fine-tune the model. This approach leveraged the language comprehension capabilities of language models.


Although these approaches have shown promising results in enhancing language model performance on specific tasks, they also present several significant limitations. These methods rely on human-annotated data and positively-rated model generations for fine-tuning pre-trained language models, which can involve considerable costs and time investments. Moreover, by relying exclusively on positively-rated data, the model's capacity to identify and address negative attributes or errors may be limited, consequently reducing its generalizability to novel and unseen data.

\subsection{Comparative Analysis of Concurrent Related Work}

In light of rapid advancements in the field, numerous concurrent works have adopted a similar philosophy, employing automated feedback to enhance language model performance~\citep{he2022rethinking,peng2023check}. While these studies jointly validate the efficacy and practicality of incorporating retrieval feedback, it is crucial to emphasize the differences between our work and these contemporary investigations. 
\citet{he2022rethinking} initially demonstrated that retrieval feedback could bolster faithfulness in chain-of-thought reasoning. However, their study is confined to commonsense reasoning tasks, and the performance improvement observed is not substantial. In contrast, our work predominantly targets knowledge-intensive NLP tasks, wherein external evidence assumes a more critical role in providing valuable feedback to augment model performance.
\citet{peng2023check} proposed utilizing ChatGPT to generate feedback based on retrieved evidence. In comparison, our research demonstrates that retrieved documents can be directly employed as feedback to refine language model outputs, significantly enhancing the efficiency of this method. Simultaneously, building on the fundamental retrieval feedback concept, we introduce two novel modules, i.e., diversifying generation outputs and ensembling initial and post-feedback answers.
In comparison to existing research, our proposed \textsc{ReFeed} methodology offers a distinctive contribution to the ongoing discourse. 
As we persist in exploring this avenue of inquiry, we foresee future studies refining these techniques, ultimately achieving even greater performance gains.

\section{Proposed Method}
\begin{figure*}[t]
    \centering
{\includegraphics[width=0.99\textwidth]{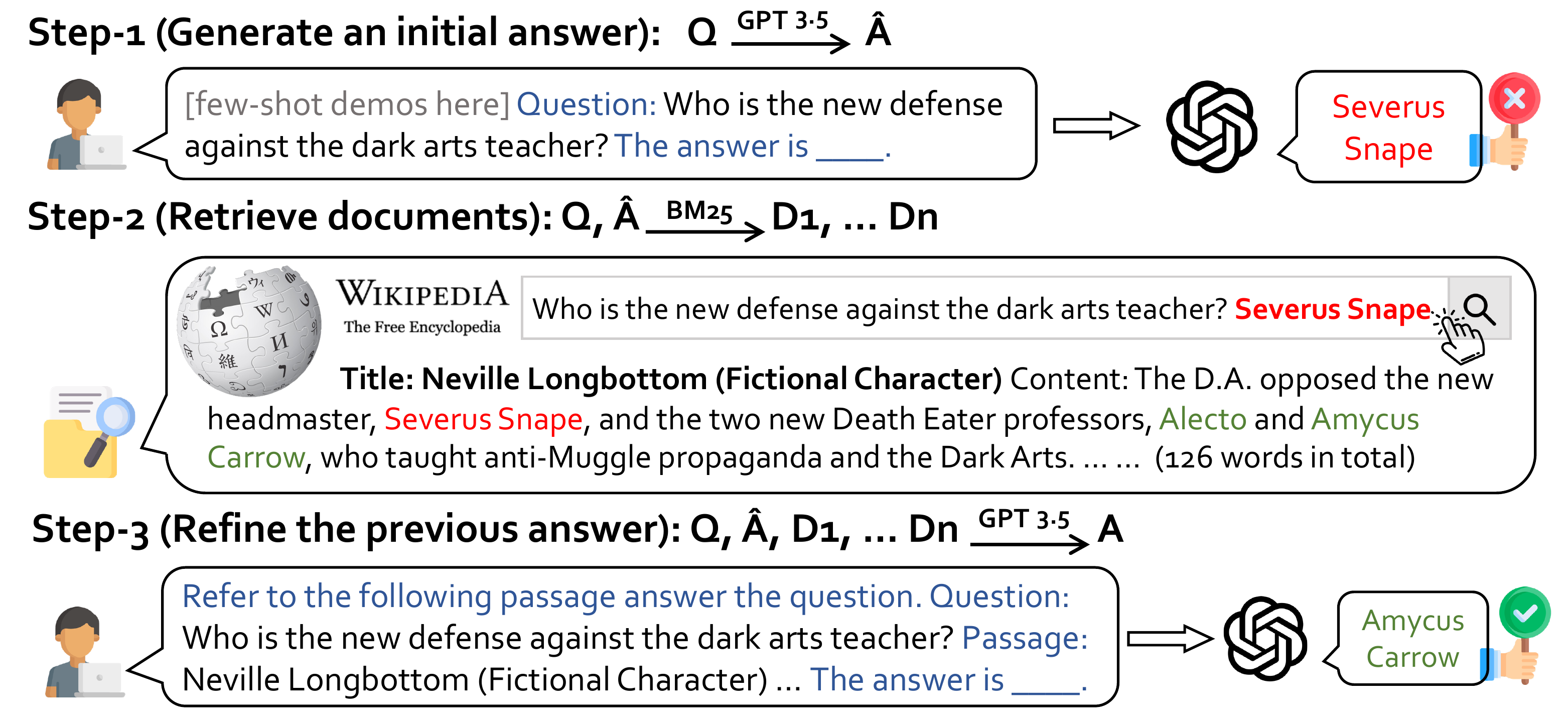}}
    \vspace{-0.1in}
    \caption{ \textsc{ReFeed} operates by initially prompting a large language model to generate an answer in response to a given query, followed by the retrieval of documents from extensive document collections. Subsequently, the pipeline refines the initial answer by incorporating the information gleaned from the retrieved documents.}
    \label{fig:framework}
    \vspace{-0.1in}
\end{figure*}

In this section, we provide an in-depth description of our innovative plug-and-play \underline{re}trieval \underline{feed}back (\textsc{ReFeed}) pipeline, specifically designed to tackle a variety of knowledge-intensive tasks ($\S$\ref{sec:main}). The pipeline operates by initially prompting a language model (e.g., InstructGPT) to generate an answer in response to a given query, followed by the retrieval of documents from extensive document collections, such as Wikipedia. Subsequently, the pipeline refines the initial answer by incorporating the information gleaned from the retrieved documents.

Besides, we introduce two novel modules based on our \textsc{ReFeed} framework. The first module aims to diversify the initial generation step, producing multiple output candidates. This enables the model to identify the most reliable answer by examining the broad range of retrieved documents. The second module employs an ensemble approach that combines language model outputs from both before and after the retrieval feedback process. This is achieved using a perplexity ranking method, which mitigates the risk of retrieval feedback inadvertently misleading the language model.

\subsection{Proposed Method: \textsc{ReFeed}}
\label{sec:main}

\textbf{Background.} Traditional large language models, such as GPT-3.5 based architectures, have primarily focused on encoding an input query and predicting the corresponding output answer~\citep{brown2020language,ouyang2022training}. In this process, the question $q$, when combined with a text prompt, serves as input to the model, which then generates the answer. This can be represented as $p(a|q, \theta)$, where $\theta$ denotes the pre-trained model parameters. In practical scenarios, the maximum a posteriori estimation (MAP) serves as the final answer, as illustrated by $\hat{a} = \mathrm{argmax}_a p(a|q, \theta)$.
However, this direct approach to eliciting answers from large language models often leads to suboptimal performance. This is because it does not fully exploit the wealth of supplementary world knowledge available to the model~\citep{levine2022standing}. To address this limitation, recent research has explored methods to improve model performance by incorporating an additional auxiliary variable, corresponding to a \textit{retrieved document} ($d$). This extension modifies the model formulation to $p(a|q) = \sum_{i} p(a|d_i, q) p(d_i|q)$, marginalizing over all possible documents.
In practice, it is infeasible to compute the sum over all possible documents ($d$) due to the vast number of potential sources. Consequently, the most common approach involves approximating the sum over $d$ using the $k$ highest ranked documents, and providing all these documents as part of the input. We assume, w.l.o.g., that these documents are $d_1, \ldots, d_k$, yielding $p(a|q) = \sum_{i=1}^k p(a|d_i, q) p(d_i|q)$. This technique is referred to as the \textit{retrieve-then-read} pipeline~\citep{lazaridou2022internet,shi2023replug}.

\begin{figure*}[t]
    \centering
{\includegraphics[width=0.99\textwidth]{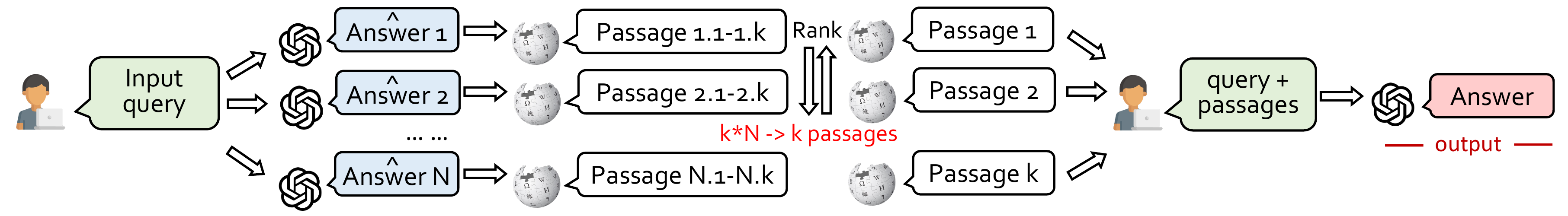}}
    \vspace{-0.05in}
    \caption{Rather than generating only one initial answer, we prompt the language model to sample multiple answers, allowing for a more comprehensive retrieval feedback based on different answers.}
    \label{fig:framework-1}
    \vspace{-0.1in}
\end{figure*}

\begin{figure*}[t]
    \centering
{\includegraphics[width=0.99\textwidth]{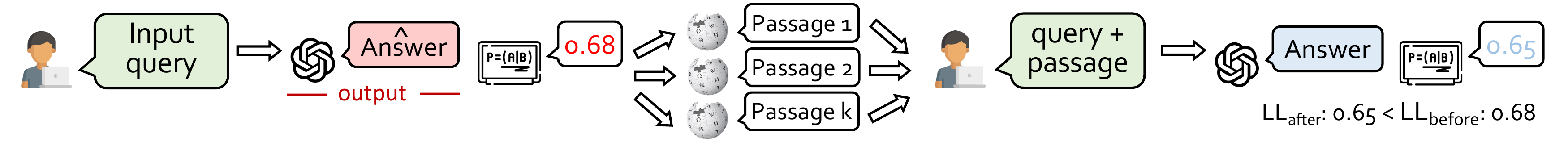}}
    \vspace{-0.05in}
    \caption{By employing an ensemble method that evaluates initial and refined answers with retrieval feedback, we enhance the assessment of answer trustworthiness, ensuring a more accurate output.}
    \label{fig:framework-2}
    \vspace{-0.1in}
\end{figure*}

\subsubsection{Basic Pipeline}
Contrary to traditional methods mentioned above, \textsc{ReFeed} is designed to offer feedback via retrieval targeted specifically to individually generated outputs. It can be formulated as $p(a|q) = \sum_{i} p(a| d_i, q, \widehat{a}) p(d_i|\widehat{a}, q)p(\widehat{a}|q)$, where $\widehat{a}$ represents the initial output, $a$ is the final output, and $d_i$ is conditioned not only on $q$ but also on $\widehat{a}$. Thus, $d_i$ is intended to provide feedback specifically on $\widehat{a}$ as the output, rather than providing general information to the query $q$. As in the case of the retrieve-and-read pipeline, we retain only the top $k = 10$ highest ranked documents: $p(a|q) = \sum_{i=1}^k p(a| d_i, q, \widehat{a}) p(d_i|\widehat{a}, q) p(\widehat{a}|q)$. 

This method enables a smooth integration of feedback to refine previous outputs in a plug-and-play fashion, eliminating the need for costly fine-tuning.
\textsc{ReFeed} takes advantage of a collection of relevant documents retrieved from an extensive textual corpus, facilitating the direct elucidation of relationships between queries and outputs. Additionally, without generating an initial output, it becomes difficult to retrieve text that supports the output due to the absence of lexical or semantic overlap with the question. Essentially, \textsc{ReFeed} functions by first prompting a large language model to produce an answer in response to a given query, followed by the retrieval of documents from external sources. The pipeline then refines the initial answer by incorporating information obtained from the retrieved documents. The three-step process is illustrated in Figure \ref{fig:framework} and outlined below.

\vspace{0.05in}
\noindent\textbf{Step 1: Generate an Initial Answer.}
In this initial step, our primary objective is to prompt a language model to generate an answer based on the given question. 
To achieve this, various decoding strategies can be employed, e.g., greedy decoding and sampling methods. In our experiments, we opted for \textit{greedy decoding} due to its simplicity and reproducibility, allowing more consistent performance across multiple runs. This step is essential for establishing a foundation upon which the following steps can build and refine the initial answer.

\vspace{0.05in}
\noindent\textbf{Step 2: Retrieve Supporting Documents.}
The second step in our pipeline involves utilizing a retrieval model (e.g., BM25) to acquire a set of  document from an extensive document collection, such as Wikipedia.
In our experiments, we retrieve top-10 documents, which offers a balanced trade-off between computational efficiency for \textit{step 3} inference and the inclusion of sufficient information.
The primary goal of this step is to identify relevant information that can corroborate or refute the relationship between the question and the initially generated answer. By extracting pertinent information from these vast sources, the pipeline can effectively leverage external knowledge to improve the accuracy and reliability of the generated response. 

\vspace{0.05in}
\noindent\textbf{Step 3: Refine the Previous Answer.}
The final step of our pipeline focuses on refining the previously generated answer by taking into account the document retrieved in \textit{step 2}. During this stage, the language model evaluates the retrieved information and adjusts the initial answer accordingly, ensuring that the response is more accurate. 
This refinement process may involve rephrasing, expanding, or even changing the answer based on the newfound knowledge. By incorporating the insights gleaned from the retrieved document, the refined answer is better equipped to address the initial query comprehensively and accurately, resulting in an improved overall performance. This step is critical for bridging the gap between the initial generation and the wealth of external knowledge, ultimately producing a high-quality, well-informed output.


\subsubsection{Enhanced Modules} 

\vspace{0.05in}
\noindent\textbf{{Diverse Answer Generation.}} Diversity in answer generation plays a pivotal role in the first step of our \textsc{ReFeed} pipeline. Rather than merely generating a single answer with the highest probability, we implement sampling methods to produce a set of potential answers. This approach fosters diversity in the generated outputs and enables a more comprehensive retrieval feedback based on diverse answers. As a result, a more refined and accurate final response is produced that effectively addresses the given question.

To elaborate, we input the question $q$ along with a text prompt into the model, which subsequently samples multiple distinct answers, denoted as $p(a_j|q, \theta)$. We then utilize the $n$ generated answers as input queries for the retrieval process, i.e., $[q, a_1], \cdots, [q, a_n]$. This stage is realized by multiple decoding passes, wherein the input query is fed into the language model with nucleus sampling.
This strategy increases the probability of obtaining a more diverse set of retrieved documents encompassing a broader spectrum of relevant information. Formally, it can be represented as $p(a|q) = \sum_{i,j} p(a| d_{i,j}, q, \widehat{a}_j) p(d_{i,j}|\widehat{a}_j, q) p(\widehat{a}_j|q)$.

Considering the limitations on the number of documents, or say context length of language model input, that can be employed in \textsc{Step 3}, we merge all retrieved documents (across different $\widehat{a}_j$), rank them based on query-document similarity scores, and retain only the top-$k$ documents for further processing. In our experiments, we use $k=10$, which offers a balanced trade-off between computational efficiency and the inclusion of diverse information. Furthermore, to account for the possibility of different queries retrieving identical documents, we perform de-duplication to ensure that the set of retrieved documents remains diverse and relevant. Lastly, when computing the final answer, we provide all $n$ generated answers as well as the aforementioned top-$k$ documents as part of the prompt.
Formally, this can be represented as 
By incorporating diversity in answer generation in \textsc{Step 1}, we effectively broaden the potential answer space, facilitating the exploration of a wider variety of possible solutions. 
$p(a|q) = \sum_{i=1}^k \sum_{j=1}^n p(a|d_{i,j}, q, \widehat{a}_j) p(d_{i,j}|\widehat{a}_j, q) p(\widehat{a}_j|q)$.


\vspace{0.1in}
\noindent\textbf{Ensembling Initial and Post-Feedback Answers.}
Retrieval feedback serves as a crucial component in obtaining relevant information to validate the accuracy of initially generated answers. Nonetheless, there may be instances where the retrieved documents inadvertently mislead the language model, causing a correct answer to be revised into an incorrect one (see examples in Figure \ref{fig:case}). To address this challenge, we introduce an ensemble technique that considers both the initial answers and the revised answers post-retrieval feedback, ultimately improving the overall generation performance.

In ensemble process, we utilize average negative log-likelihood to rank the generated answers before (i.e., $\mathrm{LL}_{\mathrm{before}}(a|q)= \frac{1}{t}\sum_{i=1}^{t}p(x_i|x_{<i}, q)$) and after incorporating retrieved documents (i.e., $\mathrm{LL}_{\mathrm{after}}(a|q)= \frac{1}{t}\sum_{i=1}^{t}p(x_i|x_{<i}, q, \widehat{a}, d)$). If the log-likelihood of an answer before retrieval feedback is higher than that after retrieval feedback, we retain the initially generated answer. On the other hand, if the log-likelihood is lower after retrieval feedback, we choose the refined answer. This strategy allows for a more informed assessment of the trustworthiness of answer before and after retrieval feedback, ensuring a more accurate final response.


\section{Experiments}

\begin{table*}
\centering
\setlength{\tabcolsep}{2.5mm}{
\begin{tabular}{l|cc|cc|cc|cc}
\toprule
\multirow{2}{*}{Models}  & \multicolumn{2}{c|}{NQ} & \multicolumn{2}{c|}{TriviaQA} & \multicolumn{2}{c|}{HotpotQA} & \multicolumn{2}{c}{WoW} \\
& EM & F1 & EM & F1 & EM & F1 & F1 & R-L \\ 
\midrule
\multicolumn{6}{l}{\textit{*close book methods without using retriever}} \\
$\mathrm{TD}$-$\mathrm{003}$~\citep{ouyang2022training} & 29.9  & 35.4 & 65.8  & 73.2  & 26.0  & 28.2 & 14.2 & 13.3 \\
GenRead~\citep{yu2023generate} & 32.5  & 42.0  & 66.2  & 73.9  & 36.4  & 39.9 & 14.7 & 13.5 \\
\midrule
\multicolumn{6}{l}{\textit{*open book methods with using retriever}} \\
Retrieve-then-Read & 31.7  & 41.2  & 61.4  & 67.4 & 35.2 & 38.0 & 14.6 & 13.4 \\
\textbf{\textsc{ReFeed} (Ours)} & \textbf{39.6} & \textbf{48.0} & \textbf{68.9} & \textbf{75.2} & \textbf{41.5} & \textbf{45.1} & \textbf{15.1} & \textbf{14.0} \\
\bottomrule
\end{tabular}}
\vspace{-0.05in}
\caption{\textsc{ReFeed} achieves SoTA performance on three 
\underline{zero-shot} knowledge intensive NLP tasks. 
The backbone model is $\mathrm{Text}$-$\mathrm{Davinci}$-$\mathrm{003}$ ($\mathrm{TD}$-$\mathrm{003}$), which is trained to follow human instructions.}
\label{tab:zero-shot}
\end{table*}


In this section, we conduct comprehensive experiments on three knowledge-intensive NLP tasks, including single-hop QA (i.e., NQ~\citep{kwiatkowski2019natural}, TriviaQA~\citep{joshi2017triviaqa}), multi-hop QA (i.e., HotpotQA~\citep{yang2018hotpotqa}) and dialogue system (i.e., WoW~\citep{dinan2019wizard}). 
In single-hop QA datasets, we employ the same splits as previous approaches \citep{karpukhin2020dense,izacard2021leveraging}. 
With regard to the HotpotQA and WoW datasets, our approach involves the usage of dataset splits provided by the KILT challenge~\citep{petroni2021kilt}. 

To thoroughly assess the performance of our model, we employ a variety of evaluation metrics, taking into consideration the professional standards established in the field. For the evaluation of open-domain QA, we use exact match (EM) and F1 score for evaluating model performance~\citep{zhu2021retrieving}. For EM score, an answer is deemed correct if its normalized form -- obtained through the normalization procedure delineated by \cite{karpukhin2020dense} -- corresponds to any acceptable answer in the provided list.
Similar to EM score, F1 score treats the prediction and ground truth as bags of tokens, and compute the average overlap between the prediction and ground truth answer.
Besides, we also incorporate Recall@K (R@K) as an intermediate evaluation metric, which is calculated as the percentage of top-K retrieved or generated documents containing the correct answer. This metric has been widely adopted in previous research~\citep{karpukhin2020dense,sachan2022questions,yu2023generate}, thereby establishing its credibility within the field.
When evaluating open-domain dialogue systems, we adhere to the guidelines set forth by the KILT benchmark~\citep{petroni2021kilt}, which recommends using a combination of F1 and Rouge-L (R-L) scores as evaluation metrics. This approach ensures a comprehensive and rigorous assessment of our model's performance, aligning with professional standards and best practices.

\begin{table*}[t]
\centering
\setlength{\tabcolsep}{2.5mm}{
\begin{tabular}{l|cc|cc|cc|cc}
\toprule
\multirow{2}{*}{Models}  & \multicolumn{2}{c|}{NQ} & \multicolumn{2}{c|}{TriviaQA} & \multicolumn{2}{c|}{HotpotQA} & \multicolumn{2}{c}{WoW} \\
& EM & F1 & EM & F1 & EM & F1 & F1 & R-L \\ 
\midrule \midrule
\multicolumn{9}{c}{\text{Backbone Language Model:  $\mathrm{Text}$-$\mathrm{Davinci}$-$\mathrm{003}$ ($\mathrm{TD}$-$\mathrm{003}$)}} \\
\midrule
\multicolumn{6}{l}{\textit{*close book methods without using retriever}} \\
$\mathrm{TD}$-$\mathrm{003}$~\citep{ouyang2022training} & 36.5 & 46.3 & 71.2 & 76.5  & 31.2 & 37.5 & 14.1 & 13.3 \\
GenRead~\citep{yu2023generate} & 38.2  & 47.3  & 71.4  & 76.8  & 36.6  & 47.5 & 14.7 & 14.1  \\
\midrule
\multicolumn{6}{l}{\textit{*open book methods with using retriever}} \\
Retrieve-then-Read & 34.3 & 45.6 & 66.5 & 70.6 & 35.2 & 46.8 & 14.5 & 13.8 \\
\textbf{\textsc{ReFeed} (Ours)}  & \textbf{40.1} & \textbf{50.0} & \textbf{71.8} & \textbf{77.2} & \textbf{41.5} & \textbf{54.2} & \textbf{15.1} & \textbf{14.3} \\
\midrule \midrule
\multicolumn{9}{c}{\text{Backbone Language Model:  $\mathrm{Code}$-$\mathrm{Davinci}$-$\mathrm{002}$ ($\mathrm{Codex}$)}} \\
\midrule
\multicolumn{6}{l}{\textit{*close book methods without using retriever}} \\
$\mathrm{Codex}$~\citep{ouyang2022training} & 41.6 & 52.8 & 73.3 & 79.2 & 32.5 & 42.8 & 16.9 & 14.7 \\
GenRead~\citep{yu2023generate} & 44.2 & 55.2 & 73.7 & 79.6 & 37.5 & 48.8 & 17.2 & 15.1 \\
\midrule
\multicolumn{6}{l}{\textit{*open book methods with using retriever}} \\
Retrieve-then-Read & 43.9 & 54.9 & 75.5 & 81.7 & 41.5 & 53.7 & 17.0 & 14.9 \\
\textbf{\textsc{ReFeed} (Ours)}  & \textbf{46.4} & \textbf{57.0} & \textbf{76.6} & \textbf{82.7} & \textbf{43.5} & \textbf{56.5} & \textbf{17.6} & \textbf{15.5} \\
\bottomrule
\end{tabular}}
\vspace{-0.05in}
\caption{\textsc{ReFeed} achieved SoTA performance on three \underline{few-shot} knowledge intensive NLP tasks.}
\label{tab:few-shot}
\end{table*}

\subsection{Backbone Language Model}

\vspace{0.03in}

\noindent$\mathrm{\mathbf{Codex}}$: OpenAI Codex, i.e., $\mathrm{code}$-$\mathrm{davinci}$-$\mathrm{002}$, a sophisticated successor to the GPT-3 model, has undergone extensive training utilizing an immense quantity of data. This data comprises not only natural language but also billions of lines of source code obtained from publicly accessible repositories, such as those found on GitHub. As a result, the Codex model boasts unparalleled proficiency in generating human-like language and understanding diverse programming languages.

\begin{table*}
\centering
\setlength{\tabcolsep}{2.3mm}{
\begin{tabular}{l|cc|cc|cc|cc}
\toprule
\multirow{2}{*}{Models}  & \multicolumn{2}{c|}{NQ} & \multicolumn{2}{c|}{TriviaQA} & \multicolumn{2}{c|}{HotpotQA} & \multicolumn{2}{c}{WoW} \\
& EM & F1 & EM & F1 & EM & F1 & F1 & R-L \\ 
\midrule
\textbf{\textsc{ReFeed} (Ours)} & \textbf{46.4} & \textbf{57.0} & \textbf{76.6} & \textbf{82.7} & \textbf{43.5} & \textbf{56.5} & \textbf{17.6} & \textbf{15.5} \\
~$\vdash$ w/o diversifying generation  & 45.1 & 56.2 & 75.9 & 82.1 & 42.1 & 54.8 & 17.0 & 14.8 \\
~$\vdash$ w/o ensemble before \& after & 45.5 & 56.5 & 76.1 & 82.4 & 42.5 & 55.3 & 17.1 & 14.9 \\
\bottomrule
\end{tabular}}
\vspace{-0.05in}
\caption{Ablation Study. Our proposed ensemble method and diversifying generation in ReFeed can improve model performance on four benchmark datasets. The backbone model is $\mathrm{Code}$-$\mathrm{Davinci}$-$\mathrm{002}$ ($\mathrm{Codex}$).}
\label{tab:abaltion-study}
\end{table*}

\noindent$\mathrm{\mathbf{Text}}$-$\mathrm{\mathbf{davinci}}$-$\mathrm{\mathbf{003}}$\textbf{:} Building on the foundation laid by previous InstructGPT models, OpenAI's $\mathrm{text}$-$\mathrm{davinci}$-$\mathrm{003}$ represents a significant advancement in the series. This cutting-edge model showcases considerable progress in multiple areas, including the ability to generate superior quality written content, an enhanced capacity to process and execute complex instructions, and an expanded capability to create coherent, long-form narratives.

After careful consideration, we ultimately decided against employing ChatGPT and GPT-4 as the foundational models for our project. The primary reason for this decision is OpenAI's announcement that both models will be subject to ongoing updates in their model parameters\footnote{https://platform.openai.com/docs/models/gpt-3-5}. These continual modifications would lead to non-reproducible experiments, potentially compromising the reliability of our research outcomes.

\subsection{Baseline Methods}

In our comparative analysis, we assess our proposed model against two distinct groups of baseline methodologies. The first group encompasses closed-book models, including InstructGPT~\citep{ouyang2022training} and GenRead~\citep{yu2023generate}, which operate without the assistance of any external supporting documents during the inference process. Each of these baseline methods adheres to a uniform input format, specifically utilizing the structure: [prompt words; question].

The second group of models adheres to a \textit{retrieve-then-read} pipeline~\citep{lazaridou2022internet,shi2023replug}, which entails a two-stage process. In the initial stage, a retriever component is employed to identify and extract a select number of relevant documents pertaining to a given question from an extensive corpus, such as Wikipedia. Subsequently, a reader component is tasked with inferring a conclusive answer based on the content gleaned from the retrieved documents. Similar to the first group, all baseline methods within this second group adhere to a standardized input format, which is defined as: [prompt words; passage; question].
As RePLUG are not a open-source model, we implemented it independently, which may result in slightly different outcomes compared to the performance reported in their respective papers.

\subsection{Experimental Analysis}

\subsubsection{Zero/Few-shot Question Answering and Dialogue Evaluation}

In the zero-shot setting, there is no training question-answer pairs and conversational input-output pairs for the models. Consequently, all models are expected to generate answers solely based on the input test question provided, without the benefit of prior training data to guide their responses.

For the purposes of our experiments, we utilized $\mathrm{text}$-$\mathrm{davinci}$-$\mathrm{003}$ as the backbone model due to its remarkable performance in zero-shot scenarios. This model excels in situations where no training question-answer pairs are available, as it is adept at generating accurate and relevant answers based solely on the input test question. As demonstrated in Table~\ref{tab:zero-shot}, our proposed \model outperforms baseline method by effectively leveraging retrieval feedback. In particular, \model exhibits a significant improvement in EM scores by +7.7 on two open-domain QA benchmarks in comparison to the original $\mathrm{text}$-$\mathrm{davinci}$-$\mathrm{003}$. We also observe a similar trend in the context of multi-hop QA tasks and dialogue systems, in which our proposed \model consistently surpasses the baseline model.

On the other hand, when juxtaposed with methods that directly retrieve or generate documents, our proposed \textsc{ReFeed} demonstrates a markedly superior performance. This can be attributed to the fact that alternative methods often struggle to retrieve relevant passages when there is an absence of lexical overlap between the query and the source text. Our proposed \textsc{ReFeed} offers a more robust and accurate solution for knowledge-intensive tasks, outpacing baseline methods across various benchmarks and experimental settings.


In the few-shot setting, as shown in Table \ref{tab:few-shot}, we observed a similar pattern to the zero-shot setting, further reinforcing the effectiveness of our proposed \model. This consistency across various settings underscores the model's versatility and adaptability, illustrating its potential to deliver superior results across a wide range of question-answering and dialogue evaluation tasks.

\begin{figure*}[t]
	\centering
     {\includegraphics[width=0.50\textwidth]{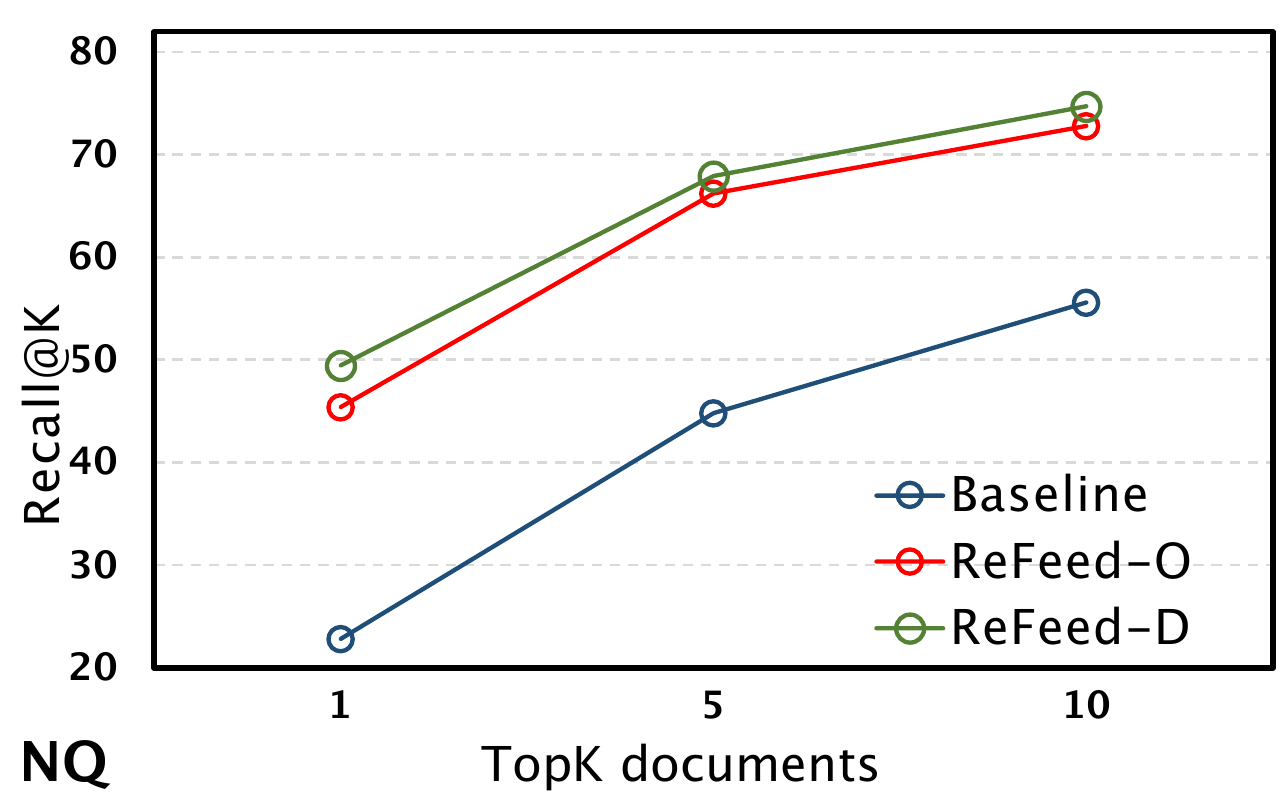}\label{fig:r@-tqa}}
	\hspace{-0.08in}
    {\includegraphics[width=0.50\textwidth]{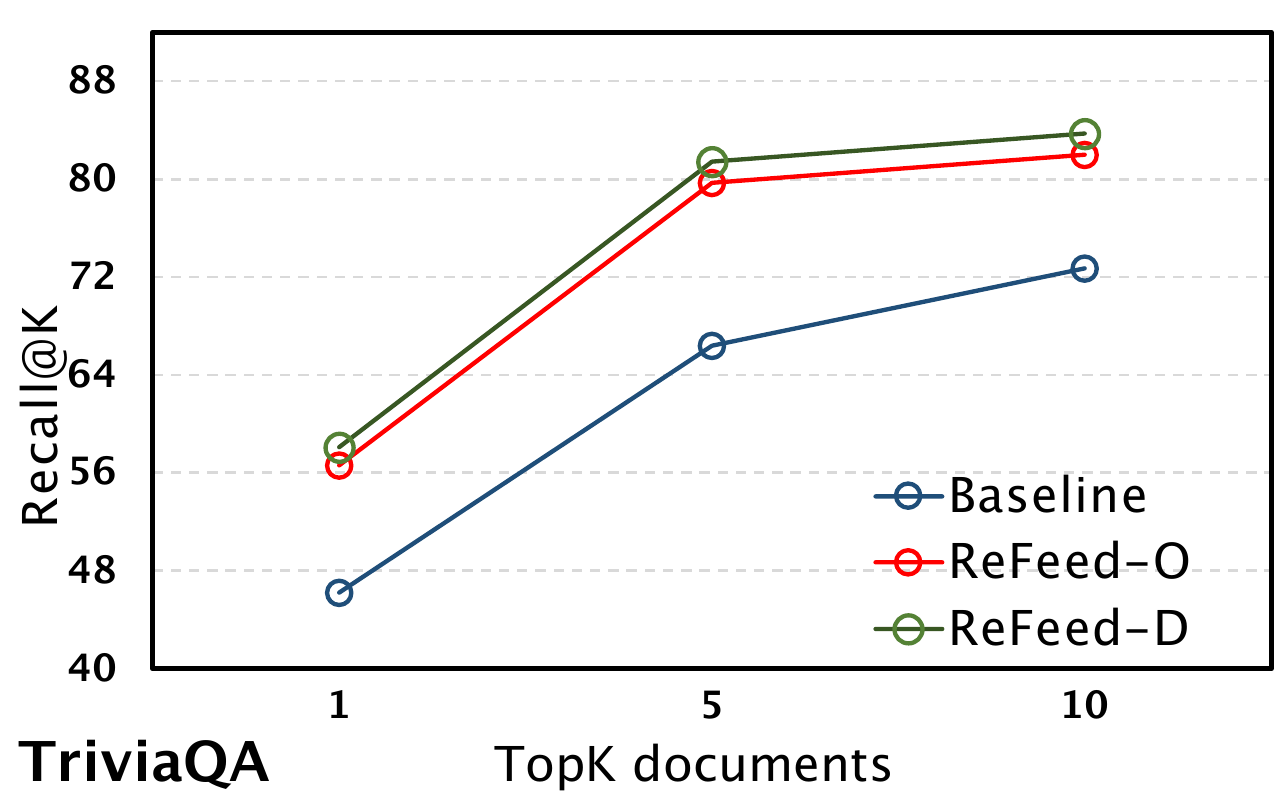}\label{fig:r@-webq}}
        \vspace{-0.2in}
	\caption{Recall@K on test sets, measured as the percentage of top-K documents that contain the answer. The ``Baseline'' refers to direct retrieval based on the input query, where the ``\textsc{ReFeed}{-O}'' represents generating only one answer, and the ``\textsc{ReFeed}{-D}'' represents diverse answer generation.}
\label{fig:retrieval-coverage}
\end{figure*}

\subsubsection{Ablation Study on Ensemble Method and Diverse Generation}

\textbf{Ensemble Method:} As shown in Table \ref{tab:abaltion-study}, it is evident that the performance of ReFeed declines by an average of 0.8 EM score across three QA datasets when the ensemble method is not employed. This finding highlights the importance of implementing an ensemble strategy both prior to and following retrieval feedback in order to bolster the model's predictive accuracy and overall effectiveness.
The ensemble method effectively utilizes the language model's inherent beliefs in conjunction with the retrieval feedback. It is worth noting that in several instances, the language model is already capable of determining the correct answer to a given question. However, the inclusion of retrieved documents may inadvertently mislead the model. The ensemble method addresses this issue by selecting the predicted answer with a \textit{higher log-probability}, thus mitigating the negative impact of the retrieved documents on the model's overall performance.

\noindent \textbf{Diverse Generation:} As shown in Table \ref{tab:abaltion-study}, the performance of ReFeed experiences a decline by an average of 1.1 EM score across three QA datasets when diverse generation is not utilized. This observation underscores the significance of incorporating diverse generation, as it can lead to multiple, distinct answers, thereby resulting in a more diverse set of documents retrieved during subsequent stages.
Diverse generation plays a crucial role in enhancing the retrieval process by diversifying the range of retrieved documents. Consequently, this increased coverage improves the overall quality and relevance of the information obtained during retrieval. As shown in Figure \ref{fig:retrieval-coverage}, the use of diverse generation through sampling techniques brings a positive improvement on the answer hit ratio, which is a consistent finding with that in the self-consistency paper~\citep{wang2023self}.
Incorporating diverse generation into the \textsc{ReFeed} framework offers several benefits, including the ability to explore a wider array of potential answers and the capacity to retrieve more comprehensive and diverse documents. With this approach, the model is better equipped to handle complex questions, ultimately leading to more accurate predictions and improved performance across various applications.

\subsubsection{Analysis on Chain-of-thought Reasoning on Multi-hop QA}

\begin{table}[t]
\vspace{-0.15in}
\setlength{\tabcolsep}{2.8mm}{\scalebox{0.95}{
\begin{tabular}{l|cc}
\toprule
\multirow{2}{*}{Models}  & \multicolumn{2}{c}{HotpotQA} \\
 & {EM} & {F1} \\ 
\midrule
No Retriever, QA Prompt        & 32.5 & 42.8  \\
No Retriever, CoT Prompt & 35.0 & 46.8 \\
Retrieve-Read with CoT Prompt & 42.1 & 54.8  \\
\textsc{ReFeed} with CoT Prompt & \textbf{44.2} & \textbf{57.4} \\
\bottomrule
\end{tabular}}}
\vspace{-0.05in}
\caption{\textsc{ReFeed} can be applied to chain-of-thought (CoT) reader as well, on multi-step reasoning task.}
\vspace{-0.1in}
\label{tab:cot-multihop}
\end{table}

Chain-of-thought reasoning entails the generation of a sequence of intermediate reasoning steps, as described in recent literature~\citep{wei2022chain}.
We demonstrate that ReFeed can be effectively integrated with chain-of-thought reasoning to address complex tasks, as opposed to merely relying on the language model to generate answers directly. 
As illustrated in Table \ref{tab:cot-multihop}, we implemented ReFeed in conjunction with chain-of-thought reasoning by generating intermediate reasoning steps prior to arriving at the final answer. Following this, we utilized the answer to retrieve documents for feedback and subsequently generated another chain-of-thought reasoning to refine the previously generated response. This approach led to a significant improvement on complex QA scenarios in the HotpotQA, when compared to employing straightforward QA prompts.

\begin{figure*}[t]
	\centering
     {\includegraphics[width=1.0\textwidth]{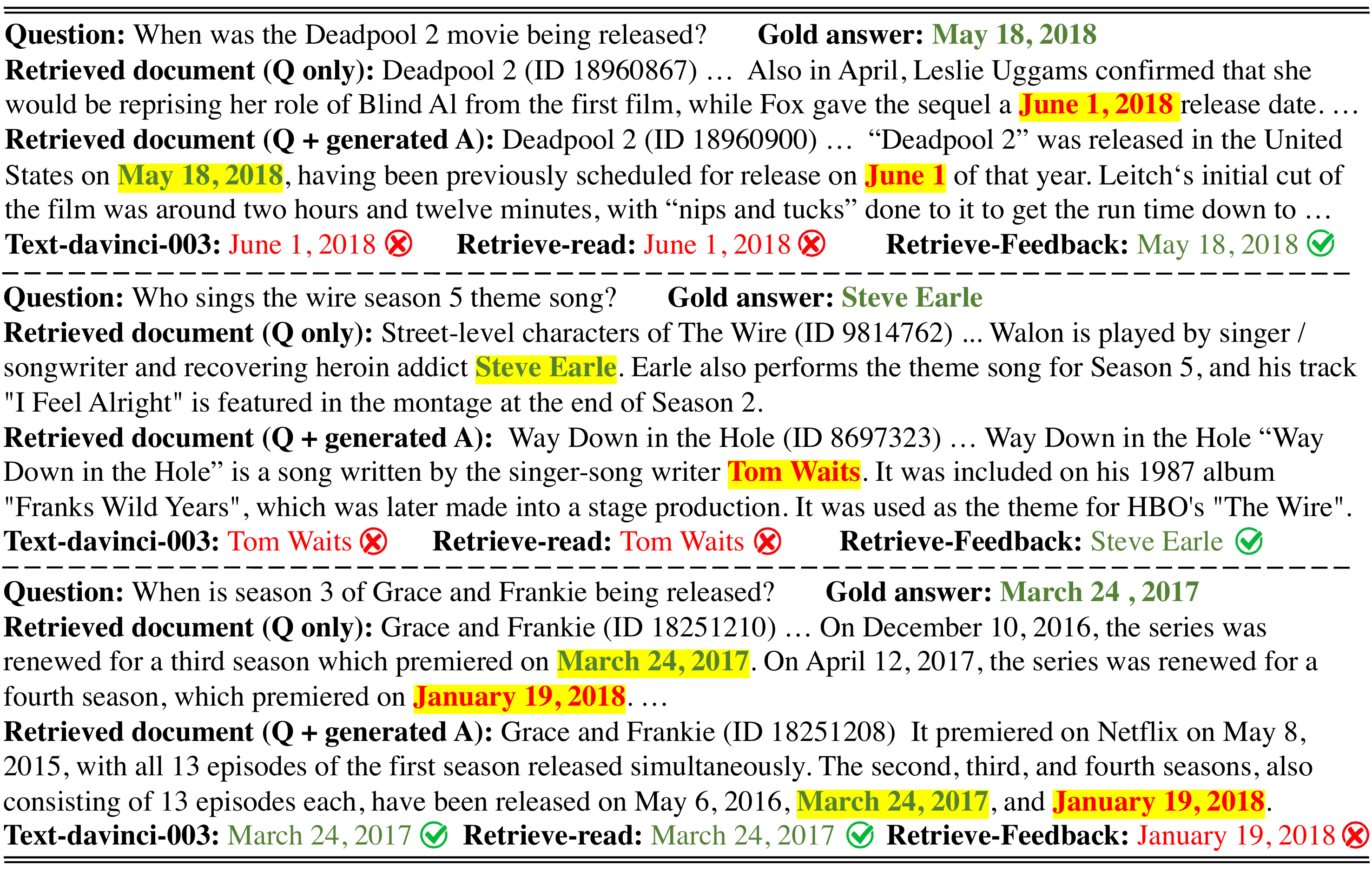}}
     \vspace{-0.25in}
	\caption{Case Studies. The first two examples illustrate how utilizing retrieval feedback leads to the generation of correct answers, while the final example demonstrates a negative outcome where the language model is misled by the retrieved document, resulting in an incorrect response.}
\label{fig:case}
\end{figure*}

To summarize, our proposed ReFeed methodology can be seamlessly integrated with chain-of-thought reasoning, thereby showcasing their complementary nature. The successful combination of ReFeed and chain-of-thought reasoning enables the model to handle more intricate tasks and exhibits its potential for tackling real-world challenges that demand complex problem-solving capabilities.

\subsubsection{Case Study}

In our case studies, we present three examples depicted in Figure \ref{fig:case} to illustrate the impact of retrieval feedback on the language model's ability to refine answers. The initial two instances showcase favorable outcomes, as the language model effectively refines answers into accurate ones by utilizing retrieval feedback. Conversely, the third instance is unfavorable, as the response is misguided by the documents after retrieval, culminating in an inaccurate response.

In the first instance, both the question-answering (QA) prompt with $\mathrm{text}$-$\mathrm{davinci}$-$\mathrm{003}$ and the retrieve-and-read model produce ``June 1, 2018'' as answer, which is erroneous. Upon investigating the retrieved document, it discloses that the film's release date was amended to ``May 18, 2018''. However, the document (Deadpool 2) is not retrieved when solely employing the query for retrieval. This occurs because, although the generated answer ``June 1, 2018'' is inaccurate, it augments the lexical overlap between the input query and candidate documents, complicating the model's capacity to pinpoint the accurate information.
In the subsequent instance, both the QA prompt with $\mathrm{text}$-$\mathrm{davinci}$-$\mathrm{003}$ and the retrieve-and-read model generate ``Tom Waits'' as the response, which is incorrect. The retrieved document elucidates that ``Tom Waits'' is the composer, rather than the vocalist. This distinction diminishes the generation likelihood of the name, facilitating the model to produce the accurate answer, ``Steve Earle'', following retrieval feedback. It is crucial to emphasize that this informative document is retrieved irrespective of whether the generated answer is employed as a component of the query for retrieval.
In the last instance, both the QA prompt with $\mathrm{text}$-$\mathrm{davinci}$-$\mathrm{003}$ and the retrieve-and-read model can generate the accurate answer. Regrettably, when incorporating the generated answer as a component of the query for retrieval, a document containing extraneous information is retrieved. This document is not retrieved in the retrieve-then-read pipeline. Alas, this document misdirects the language model, ultimately yielding an inaccurate answer.

\section{Conclusion}
In conclusion, this paper presents a novel pipeline, \textsc{ReFeed}, designed to improve large language models' performance in a plug-and-play framework, effectively addressing the challenges arising from knowledge-intensive tasks. By employing a retrieval method to provide automatic feedback on generated outputs and integrating this feedback to refine the outputs without the need for expensive fine-tuning, \textsc{ReFeed} offers a practical and efficient solution.
We introduce two innovative modules within the \textsc{ReFeed} pipeline: diverse answer generation and an ensemble approach. These two modules further enhance \textsc{ReFeed} to produce more reliable and accurate answers by considering a wider array of retrieved documents and mitigating the risk of misleading retrieval feedback.
Our extensive experiments on four challenging knowledge-intensive benchmarks demonstrate the effectiveness of \textsc{ReFeed} in achieving state-of-the-art performance under the few-shot setting. We believe by continuing to refine and optimize the \textsc{ReFeed} pipeline, we can unlock its full potential and expand its applicability across a diverse range of scenarios and applications.

\bibliography{reference}
\bibliographystyle{acl_natbib}


\end{document}